\documentclass[conference]{IEEEtran}
\IEEEoverridecommandlockouts
\usepackage{amsmath,amssymb,amsfonts}
\usepackage{algorithmic}
\usepackage{graphicx}
\usepackage{textcomp}
\usepackage{xcolor}
\usepackage{booktabs}
\usepackage{verbatim}
\usepackage{hyperref}
\usepackage{url}
\newcommand{\projectname}{CANOPIES}
\newcommand{\projecturl}{\scriptsize\url{https://canopies.inf.uniroma3.it/consortium}}
\newcommand{\coderepo}{\scriptsize \url{https://fra-tsuna.github.io/llm-to-symbolic-planner/}}
\def\BibTeX{{\rm B\kern-.05em{\sc i\kern-.025em b}\kern-.08em
    T\kern-.1667em\lower.7ex\hbox{E}\kern-.125emX}}
\begin{document}

\title{Defining and Monitoring Complex Robot Activities
via LLMs and Symbolic Reasoning}

\author{%
  \IEEEauthorblockN{Francesco Argenziano, Elena Umili, Francesco Leotta, and Daniele Nardi}
  \IEEEauthorblockA{\textit{Department of Computer, Automation and Management Engineering}\\
                    Sapienza University of Rome, Rome, Italy\\
                    \{\texttt{{argenziano, umili, leotta, nardi\}@diag.uniroma1.it}}}
}

\maketitle

\begin{abstract}
Recent years have witnessed a growing interest in automating labor-intensive and complex activities, i.e., those consisting of multiple atomic tasks, by deploying robots in dynamic and unpredictable environments such as industrial and agricultural settings. A key characteristic of these contexts is that activities are not predefined: while they involve a limited set of possible tasks, their combinations may vary depending on the situation. Moreover, despite recent advances in robotics, the ability for humans to monitor the progress of high-level activities - in terms of past, present, and future actions - remains fundamental to ensure the correct execution of safety-critical processes. In this paper, we introduce a general architecture that integrates Large Language Models (LLMs) with automated planning, enabling humans to specify high-level activities (also referred to as processes) using natural language, and to monitor their execution by querying a robot. We also present an implementation of this architecture using state-of-the-art components and quantitatively evaluate the approach in a real-world precision agriculture scenario.
\end{abstract}


\section{INTRODUCTION}
\label{sec:intro}

In recent years, there has been a significant increase in the interest and demand for automating complex and labor-intensive activities through the deployment of robotic systems. These activities, often encountered in industrial and agricultural domains, are typically composed of multiple, smaller atomic subtasksthat must be coordinated to achieve a larger goal. What makes these environments particularly challenging is their dynamic and unpredictable nature: the specific sequence and combination of tasks required can change based on real-time conditions, external events, or evolving objectives. Importantly, while the range of possible jobs is usually limited and known in advance, the structure and flow of the overall activity are not fixed and therefore must be adapted on the fly.\looseness-1

In such contexts, human operators must maintain a clear understanding of ongoing high-level activities, both to ensure correctness and to make timely decisions. This includes being aware of what the system has done (past), what it is currently doing (present), and what it intends to do next (future). However, this kind of situational awareness is difficult to maintain in the absence of mechanisms for querying the system in a human-friendly way.\looseness-1

High-level activities (in the following the term \emph{processes} will be also used for brevity) can be described in different ways, usually employing formal methods and languages such as Petri Nets or temporal logics~\cite{dumas2018fundamentals}. They can be imperative (i.e., only what is explicitly specified is allowed) or declarative (i.e., what is not forbidden is allowed). Their definition is usually transparent and independent from the actual functionalities provided by the participating actors. As a consequence, in order to actually enact them, they must be bound to a sequence of executable actions that can be obtained, for example, by using automated synthesis techniques (e.g., automated planning~\cite{ghallab2004automated}).\looseness-1

In addition, when collaboration involves humans and robots, the way information is conveyed is fundamental from the point of view of usability, with speech being a natural way for a human to convey information with a robot~\cite{kaszuba2021preliminary}. Recently, the world of Natural Language Processing (NLP) has been revolutionized by the emergence of Large Language Models (LLMs), advanced computational models capable of operating on and generating sentences in natural language~\cite{chang2023survey}.\looseness-1

State-of-the-art large pre-trained language models (PLMs) such as BERT~\cite{devlin2018bert}, GPT~\cite{openai2024gpt4technicalreport, openai2024gpt4ocard} and LLaMA~\cite{touvron2023llama} are not domain-specific, and their output can be wrong or ambiguous (this phenomenon is referred to as \emph{hallucination}) when applied to specialized applications. Developing a domain-specific model requires a proper dataset and significant computing power for LLM training (or fine-tuning), which may not always be available. In this case, the \emph{in-context learning} ability of LLMs paired with the \emph{prompt engineering} approach, becomes useful, enabling zero-shot or few-shot learning without additional training data~\cite{dai2022can,zhao2023explainability}. Users design specific prompt text, sometimes with few examples (\emph{few-shot prompting}), to guide an LLM in generating desired responses.\looseness-1

In this paper, we introduce a framework that enables users to specify high-level activities in natural language, making it accessible to non-experts. Additionally, it allows users to interact with the system during execution, asking questions about task progress, future steps, or explanations of behavior. This improves transparency, usability, and trust in robotic systems operating in open and evolving environments. In particular, we aim at the following design goals:
\begin{itemize}
\item Not relying on a fixed set of processes. Conversely, the objective is to let humans define them on the fly, according to the needs encountered on the field, and to be aware of what the robot is going to do in order to satisfy the expressed goals;
\item Constraining processes to tasks actually executable by the robot on the environment. This means modeling the environment and possible actions formally;
\item Integrating process monitoring by human with other information related to the state of the world;
\item Relying on natural language alone for Human-Robot communication.
\end{itemize}

We validate our approach with a full implementation of the proposed architecture using state-of-the-art LLMs and planning systems. The solution is evaluated in a real-world precision agriculture scenario, where it demonstrates the effectiveness and adaptability of the approach in supporting both autonomous task execution and intuitive human-robot interaction.\looseness-1

The paper aims to show how integrating symbolic reasoning with modern LLMs can benefit users by \emph{(i)} providing precise solutions using symbolic reasoning, \emph{(ii)} simplifying interaction by shielding users from the complexity of symbolic reasoning formalisms, \emph{(iii)} leveraging LLMs for natural language interaction, and \emph{(iv)} relieve the LLM of performing planning tasks, which often leads to hallucinations.\looseness-1

The paper is structured as follows. Section~\ref{sec:related} introduces related works, contextualizing the work with respect to the state of the art. Section~\ref{sec:architecture} describes the proposed approach from the conceptual and implementation point of view. Section~\ref{sec:evaluation} evaluates the approach with a quantitative analysis in a challenging precision agriculture scenario. Section~\ref{sec:conclusions} concludes the paper with discussion and future works.
\section{RELATED WORKS}
\label{sec:related}

\subsubsection*{\textbf{Behavioral modeling and process formalisms}}
Early work modeled robot and human behaviors with Finite State Automata (FSA) \cite{fsa_robot_programming}. FSAs are simple and intuitive but scale poorly: complex behaviors require exploding state spaces that are hard to design, maintain, and verify. To increase expressivity, researchers adopted richer formalisms such as BDI (Belief-Desire-Intention) architectures \cite{rao1997modeling}, Petri nets \cite{ziparo2011petri}, and process-modeling techniques from business-process research \cite{dumas2018fundamentals,process_modelling}. BDI systems provide flexible, plan-selection semantics (intentions chosen from a plan library) that avoid committing designers to fixed action sequences, but they typically lack integrated automated planning and formal validation tooling. Petri nets and related imperative models reduce state explosion compared to FSAs but still require enumerating allowed behaviors a priori, which can be restrictive in open or human-interactive settings. Declarative process formalisms based on temporal logics (e.g., LTL) shift the specification burden from enumerating traces to stating constraints, enabling more compact specifications and supporting runtime verification; several robotics works exploit these properties for safety and constraint enforcement \cite{LTL,baran2021ros}. Our approach leverages the compactness and verifiability of temporal-logic specifications while combining them with synthesis to produce enactable plans rather than only enforcing constraints at runtime.\looseness-1

\subsubsection*{\textbf{Automated planning and synthesis for robotics and smart environments}}
Automated (task) planning provides a principled way to derive action sequences from symbolic action models \cite{karpas2020automated}. Recent work applies synthesis techniques and automated planners to robotics and smart environments to generate controllers or mission plans from high-level requirements \cite{guo2023recent,de2023digital,monti2023suitability}. These methods enable correctness-oriented behavior generation and can incorporate domain models that reflect the robot’s actual capabilities. However, a common shortcoming in many prior systems is the weak coupling between specification, synthesis, and executable enactment: some approaches stop at verification or controller synthesis that is not directly mapped to robot primitives. In contrast, we define an action theory that grounds the synthesized plan in the robot’s actionable repertoire and integrate LTL constraints with classical planning to obtain execution-ready plans.\looseness-1

\subsubsection*{\textbf{Monitoring, execution-awareness, and human–robot collaboration}}
Much of the literature emphasizes plan generation and execution but tends to treat monitoring and human interaction as secondary concerns. Works that focus on collaboration often consider multi-robot scenarios more than human-robot interaction (HRI), and many models emphasize execution semantics at the expense of continuous monitoring. Petri nets and imperative models, while useful for coordination, do not naturally support flexible runtime adaptation or lightweight monitoring of human intent. Declarative specifications (LTL) and synthesis approaches open the door to runtime checks and recovery, but prior work has generally either enforced constraints reactively \cite{baran2021ros} or generated multi-agent plans without an explicit, interactive monitoring loop. Our contribution emphasizes explicit, human-to-robot communication of high-level intentions and continuous monitoring during enactment: humans provide high-level goals, the system synthesizes temporally constrained, enactable plans, and monitoring keeps the human informed about progress and deviations.

\subsubsection*{\textbf{LLMs and hybrid symbolic–neural approaches}}
LLMs have recently been applied to HRI for natural-language understanding, dialogue, and flexible question answering \cite{zhang2023large}. Prompting and Retrieval-Augmented Generation (RAG) make LLMs powerful for handling out-of-domain queries without retraining \cite{lewis2020retrieval}. Nevertheless, LLMs’ deep reasoning and guarantee properties remain debated, so several recent efforts combine LLMs with symbolic planners or synthesizers to get the best of both worlds. For example, hybrid pipelines employ LLMs for intent extraction or domain/problem construction and delegate formal reasoning to planners \cite{liu2024towards}. Our design follows this hybrid philosophy: we use LLMs for NLP and shallow reasoning (intent interpretation and dialogue) but rely on automated synthesis and temporal-logic based planning to produce correct, executable plans and to drive monitoring. Compared to prior hybrid efforts, we emphasize explicit (rather than implicit) human instruction, support higher-level goals with temporal constraints, and tightly integrate monitoring during activity execution.\\

In sum, prior work provides powerful but fragmented tools: expressive specification languages and verification (declarative/LTL), practical plan generation (automated synthesis), and flexible natural-language interfaces (LLMs). However, gaps remain in combining these capabilities into a single pipeline that \textit{(i)} produces enactable plans grounded in robot capabilities, \textit{(ii)} supports explicit human-to-robot goaling with temporal constraints, and \textit{(iii)} continuously monitors execution and reports progress to the human. Our approach explicitly targets these gaps by integrating LTL constraints, automated planning for enactability, and an LLM-assisted interface for human instructions and monitoring.
\section{PROPOSED APPROACH}
\label{sec:architecture}

\begin{figure}[t!]
      \centering
      \includegraphics[scale=0.255]{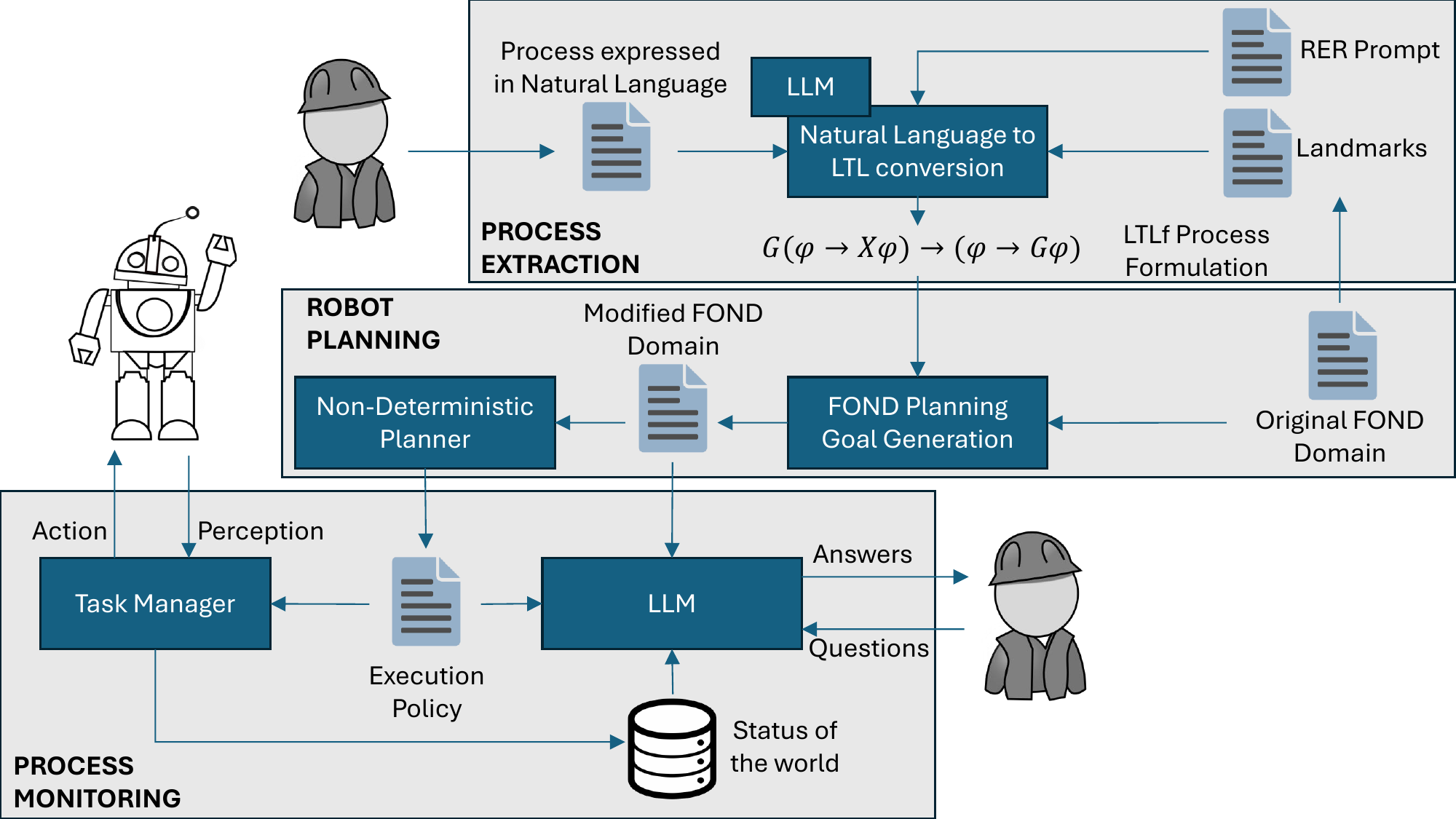}
      \caption{Architecture of the proposed solution}
      \label{fig:architecture}
      \vspace{-15pt}
   \end{figure}

Figure \ref{fig:architecture} shows the the framework proposed in this paper. The implementation is available as an open source tool\footnote{\coderepo{}}. The general idea is to translate natural language descriptions of high-level activities provided by a human operator into a plan executable in the environment whose execution can be monitored by the human asking natural language questions. To generate the plan, the system relies on a Planning Domain Definition Language (PDDL) file that describes the actions available in the environment in terms of pre- and post-conditions. By specifying a goal, expressed in terms of predicates on the environment, a planner is able to define a plan that can be executed and queried. To specify the goal, we turn a high-level activity expressed in natural language (e.g., through a microphone) into a logic formula. Automated planners usually take this goal formula in the form of a conjunction of truth predicates. We instead employ Linear Temporal Logic interpreted over finite traces (LTLf) \cite{de2013linear} as an intermediate representation formalism, which simplifies the translation from natural language. Once a specific plan is computed, the human can ask questions during its execution about the status of the activity and its next steps.\looseness-1

Let us delve into the details of the architecture depicted in Figure~\ref{fig:architecture}.
The system takes as input an utterance, describing a high-level activity, which is converted to a string, e.g., through any speech-to-text module. This string is given as input to a Lang2LTL~\cite{liu2023grounding} module that converts the string into an LTLf formula using predicates taken from the planning domain (see Section~\ref{sec:speechtoltl}).\looseness-1

The LTLf formula and the planning domain are then fed into the FOND4LTLf~\cite{de2021fond4ltl} module, which produces an augmented planning domain, which extends the original non-deterministic domain by adding predicates and actions, thus allowing to support LTLf formula as goal of a planning problem. The new domain and problem are finally given as input to a non-deterministic planner that produces a policy that is able to achieve the task (see Section~\ref{sec:planning}). Noticeably, the planner belongs to the FOND (Fully Observable Non-Deterministic) class, thus allowing the user to check the status of the system after the execution of a specific action in the plan.\looseness-1

At this point, an LLM is used to turn the plan into a text and to answer questions, in natural language, about the execution of the plan (see Section~\ref{querying}).  The plan (or execution policy) is also fed into the Task Manager module, which is in charge of executing the single actions and updating a knowledge base with the status of the world, which is used as part of the prompt for the dialog. As LLMs, we used GPT-3.5 for process extraction (Section \ref{sec:processextraction}) and GPT-4o \cite{openai2024gpt4technicalreport} for process monitoring  (Section \ref{llmanswers}).\looseness-1

To illustrate our approach, and later in Section~\ref{sec:evaluation}, we will refer to a real world precision agriculture scenario defined in the \projectname{} project\footnote{\projecturl{}}. \projectname{} aims to develop a novel collaborative human-robot paradigm in the field of precision agriculture for permanent crops, where farmworkers can efficiently work together with teams of robots to perform agronomic interventions, like harvesting or pruning table-grape vineyards. Provided examples are self-explainable, but for the sake of readability the scenario defined by the project includes two types of robots. A \emph{harvesting robot}, which is in charge of all the operations on the grapes, including quality estimation and harvesting, and a \emph{support robot}, which takes care or moving harvested bunches to the warehouse. The field is organized in lines following the typical trellis layout.\looseness-1

\subsection{Process Extraction}
\label{sec:speechtoltl}

Transforming natural language sentences into a formal logic representation, such as Linear Temporal Logic (LTL), presents a significant challenge. After thoroughly reviewing systems documented in the literature, we performed specific tests on NL2LTL~\cite{fuggitti2023nl2ltl} and Lang2LTL~\cite{liu2023grounding}. Ultimately, we chose Lang2LTL for our task due to its modular design and its extensive training, which allows it to recognize a wide variety of formula patterns. These characteristics make possible to apply Lang2LTL to a different problem with minimal fine-tuning data, unlike~\cite{fuggitti2023nl2ltl}, which would require much more examples.\looseness-1

Lang2LTL does have some limitations though. First, it still depends on specific patterns and struggles to generalize to unseen LTL formulas, similar to NL2LTL~\cite{fuggitti2023nl2ltl}. More critically, Lang2LTL focuses on translating \textit{navigation commands}, meaning the symbols in the formulas refer to specific waypoints within the environment that must be visited, avoided, or revisited in a particular order. This aspect restricts the range of expressions that the tool can handle.\looseness-1

In this work, we challenged Lang2LTL by applying it to sentences referring to \textit{locations}, \textit{conditions}, and \textit{actions}, inspired by \cite{robotic_patterns}. More formally, we define the alphabet of possible LTL formulas as the set of symbols \( P = PL \cup PC \cup PA \). 
Specifically, given a finite set of locations \( L = \{l_1, l_2, ..., l_n\} \) and robots \( R = \{r_1, r_2, ..., r_n\} \), we define \( PL = \{r_x \text{ in } l_y | r_x \in R \text{ and } l_y \in L\} \) as the set of location symbols, where each symbol indicates that a robot \( r_x \) is in a specific location \( l_y \) of the environment at the present moment.\looseness-1
Given a finite set of environmental conditions \( C = \{c_1, c_2, ..., c_m\} \), we define \( PC = \{s_1, s_2, ..., s_m\} \) where \( s_i \) is true if and only if condition \( c_i \) holds at the present moment.
Finally, the set of action symbols \( PA \) given a finite set of actions \( A = \{a_1, a_2, ..., a_m\} \) that the robots can perform, we define \( PA = \{r_x \text{ executes } a_y | r_x \in R \text{ and } a_y \in A\} \), i.e., \( r_x \) executes \( a_y \) if and only if robot \( r_x \) performs action \( a_y \) at the present moment.\looseness-1

Using this expanded set of symbols, we can specify, for example, that a robot must never perform a certain action in a particular location, or that it should execute certain tasks only when a specific environmental condition is satisfied. These types of specifications align more closely, for example, with the tasks we aim to express within the \projectname{} domain.\looseness-1

Our adapted Lang2LTL operates through three distinct phases, which we summarize briefly below:
\subsubsection{Referring Expressions Recognition (RER)} The first step is to identify expressions within the sentence that correspond to concepts that will serve as symbols in the formula. Specifically, this task is prompted to GPT-3.5, along with a set of example inputs with the corresponding outputs. An example of this first phase follows:
\begin{itemize}
\item {\textbf{Input}: A natural language sentence such as, ``You cannot call the support robot without visiting line $1$ right before, and you cannot visit line $1$ without calling the support robot right after that".}
\item {\textbf{Output}: A set of sub-phrases that refer to key concepts (\textit{referring expressions}), such as {``call the support robot", ``line $1$", ``calling the support robot"}.}
\end{itemize}

\subsubsection{Symbolic Translation} In this phase, the process description is analyzed to extract the temporal relationships between the referring expressions, producing an LTL formula. The process begins by replacing each referring expression with a placeholder symbol (such as $a$, $b$, $c$, etc.). The authors of Lang2LTL retrained GPT models specifically for the symbolic translation task using pairs of symbolic sentences and corresponding LTL formulas. These newly trained models are freely available to the community. Substituting referring expressions with symbols before training makes this module largely domain-independent. For example:
\begin{itemize}
\item {\textbf{Input}: A ``symbolic" sentence where the referring expressions are replaced by placeholder symbols (e.g., "You cannot $a$ without visiting $b$ right before; and you cannot visit $b$ without $a$ right after that").}
\item{\textbf{Output}: An ungrounded LTL formula that uses placeholders as symbols (e.g., $G (b \iff X a)$).\footnote{The syntax used in this paper~\cite{de2021compositional} includes: X – Next, WX – Weak Next, U – Until, F – Eventually, G – Globally.}}
\end{itemize}

\subsubsection{Symbol Grounding} The final step is to associate placeholder symbols with the symbols of the specific domain of interest. This is achieved using ``landmarks", i.e., couples consisting of an identifier and associated data. The identifier represents a domain symbol that can be linked to a placeholder symbol, while the data include additional fields used for grounding. In this phase, GPT is employed once again, mapping the referring expressions identified in Phase 1, along with the information related to landmarks, into the latent space of the model. Each referring expression is then matched to the landmark that minimizes the cosine similarity between their respective latent points. This phase involves more than a simple substitution, as the expressions from Phase 1 do not necessarily correspond to predicates in the planning domain. Thus, the LLM is tasked with associating these expressions with meaningful terms for the specific planning domain.
\begin{itemize}
\item{\textbf{Input}: A generic LTL formula (e.g., \(G(b\iff Xa)\)).}
\item{\textbf{Output}: An LTL formula grounded into the domain (e.g., \(G (robot\_at\_loc\_l\textit{1} \iff X (call\_support))\)).}
\end{itemize}

To recap, with respect to the classical Lang2LTL, we: \begin{itemize} \item extended the prompt file used in Phase 1 to accommodate the identification of concepts related not only to spatial locations (as the original tool focuses on navigation tasks~\cite{liu2023grounding}), but also to conditions and actions. \item created a custom landmarks file for our PDDL domain in Phase 3, using domain symbols (predicates, constants, etc.) as identifiers, and providing concise natural language descriptions of each symbol's meaning. \end{itemize}

\subsection{Planning}
\label{sec:planning}

After applying Lang2LTL as described in Section~\ref{sec:speechtoltl}, we have an LTLf formula that specifies the high-level activity the human wants the robot to perform.

To develop a planning procedure that incorporates this formula as the agent’s ultimate goal, a non-deterministic PDDL formalization of the specific domain is needed. In the \projectname{} domain, for example, the physical space is discretized into several meaningful locations. In these locations, there are grapes which can be found in various states: \textit{ripe}, \textit{unripe}, or \textit{unknown} (if an error occurs). Grapes may or may not be harvested depending on their condition. Additionally, the domain includes a support robot that assists when the main logistic robot’s harvesting box is full. We defined key actions that enable the robot’s primary functions, such as moving through the vineyard, harvesting grapes, calling the support robot, and more. The non-determinism arises from the varying states of the grapes.\looseness-1

The problem file in the PDDL description outlines the objects in the domain and specifies the initial state. However, the goal is not predefined; it changes dynamically, based on the LTLf formula we want to satisfy. This goal is generated just before providing the PDDL to the planner.\looseness-1

Once we have formalized the domain and problem, the next step is to feed these files into a planner to generate a policy. However, this introduces two challenges: \emph{(i)} the actions are non-deterministic; \emph{(ii)} the goal is not a simple conjunction of fluents\footnote{A fluent is a function that changes value over time. A predicate is a type of fluent with Boolean values.}, as it is typically the case in automated planning, but a complex LTLf formula over these fluents. \looseness-1

At this point, we translate the LTLf formula into a goal suitable for our planning domain by using \textit{FOND4LTLf}~\cite{de2021fond4ltl}, a tool that compiles Fully Observable Non-Deterministic (FOND) planning problems with temporally extended goals (like LTLf in our case) into classical FOND planning problems. The output of this process is an extended version of our original domain and problem files, where the satisfaction of the LTLf formula is encapsulated, allowing us to use a standard conjunction of fluents as the goal.\looseness-1

Finally, we use the \textit{Planner for Relevant Policies} (PRP) tool~\cite{muise-icaps-14} to obtain a policy for our planning task that accounts for non-determinism and ensures the satisfaction of the LTLf formula, eventually achieving the desired goal.\looseness-1

\subsection{Process Monitoring}
\label{querying}

Once the policy is obtained, the robot follows it to achieve its goal. In the context of process monitoring, it is crucial that we understand what the robot is doing and why, in the most natural and accessible way possible. While it is feasible to track the robot's state within the problem's state space and infer its next steps based on the unsatisfied goals, a simpler and more intuitive approach is preferred—especially for a farmer working alongside the robot.\looseness-1

To facilitate this, we designed a Q\&A mechanism integrated with an LLM, specifically GPT-4o~\cite{openai2024gpt4ocard}. This system allows the robot to explain its actions, reasoning, and future steps in a natural, conversational manner. The pipeline operates as follows:

\begin{enumerate} \item The human asks the robot a question (e.g., ``\textit{What is your next action and why?}"). \item A query is issued to GPT-4o. It contains the question, the policy generated by the planner and the PDDL files. \item The model’s response is delivered to the user. 
\end{enumerate}

With this procedure, we can monitor robot actions, from the human perspective.
\subsection{Beyond Process Monitoring}
\label{sec:beyond}

Monitoring robots can be intended, in general, at various levels of granularity. While the approach presented in this paper focuses on monitoring the process at an high level of abstraction, LLMs may also support it at other levels.\looseness-1

As outlined in previous sections, the natural language specifications are ultimately grounded in a plan comprising a sequence of basic actions available within the PDDL domain. These actions are then translated by the task manager to invoke services developed within the robotic platform ecosystem that extract information and/or execute actions.\looseness-1

For instance, in the \projectname{} domain, a set of robot actions contributes to various aspects of environmental monitoring, such as grape quality assessment, human action recognition, and navigation within the environment. These functionalities are integrated into the monitoring module through the task manager, which maintains a knowledge base of acquired environmental information. If certain information is unavailable, the task manager calls the appropriate service to obtain it. The knowledge base is queried via the LLM, similar to how the status of process execution is monitored.\looseness-1

The actions can be either very fine-grained or considered as activities composed of multiple actions. In cases where specifying details using natural language becomes impractical due to excessive granularity, the task manager can map a single PDDL action to a more complex activity. This approach enables the definition of \emph{hierarchical} processes~\cite{dumas2018fundamentals} with minimal effort required from the human user.\looseness-1
\section{EVALUATION}
\label{sec:evaluation}

In this section, we evaluate the performance of the proposed methodology by separately assessing the process extraction and process monitoring phases. For a quantitative analysis, we generated a \emph{synthetic} dataset while, for a qualitative analysis, the \projectname{} simulator was employed.

\subsection{Evaluating Process Extraction}
\label{sec:processextraction}

As shown in Figure~\ref{fig:architecture}, the FOND domain is a fundamental input to this phase. For \projectname{}, it features 23 symbols: 4 for locations, 13 for conditions, and 6 for actions. For each symbol (that corresponds to a fluent in the FOND domain), we defined a brief natural language description, which is included in the landmark file used by Lang2LTL to associate the symbols with referring expressions during grounding.

We created a dataset including over 500 natural language sentences relevant to our domain. Approximately 200 sentences express navigation patterns (such as \textit{Visit}, \textit{Sequenced Visit}, \textit{Ordered Visit}, \textit{Strict Ordered Visit}, and \textit{Global Avoidance}) using only location symbols. The remaining 300 sentences refer to generic tasks using all symbols in \(P\), corresponding to patterns like \textit{Bound Delay}, \textit{Delayed Reaction}, \textit{Prompt Reaction}, \textit{Wait}, \textit{Past Avoidance}, and \textit{Future Avoidance}. For a detailed explanation of the patterns used, we point the reader to \cite{robotic_patterns}.\looseness-1

To simplify the creation of the dataset and minimize bias, we used GPT-3.5. For each pattern, we fed GPT with a description of the pattern, an overview of the available symbols, and three example pairs (sentence, LTL formula) to guide the generation of additional data. The generated data was double-checked to ensure quality and relevance.\looseness-1

Accuracy of process extraction is measured as the percentage of LTLf formulas correctly extracted from text.

\subsection{Evaluating Process Monitoring}\label{llmanswers}

Quantifying the ability of a system to monitor a process is a complex challenge. Our objective, in particular, is to assess whether an LLM-driven embodied agent, operating in a collaborative scenario with humans and robots, can provide explanations regarding its actions, and whether these explanations are informative and aligned with the agent execution and intents.\looseness-1 

We focused on three primary evaluation scenarios: assessing the ability to monitor what the system is \textit{currently} doing, what it has done \textit{in the past up to the current state}, and what it will do \textit{in the future starting from this state}. For each scenario, we defined six questions to input into the LLM (along with the policy and PDDL files, as described in Section~\ref{querying}) to elicit responses from the system. During the evaluation, a sentence is uniformly sampled from the corresponding set based on the scenario being considered. This approach minimizes bias from specific input patterns that may influence the responses.

For each scenario, the quality of the robot's responses is evaluated by comparing the fluents predicted in the response with those that are actually true in the various states of the plan. To extract the predicted fluents we convert the natural language answer provided by the robot into a set of fluents. To this purpose, we employed another instance of GPT-4o, prompting it to function as a $FluentExtractor(\cdot)$. This function takes the following inputs: \emph{(i)} a natural language sentence ss that describes a state (e.g., \textit{I am currently in the initial location}); \emph{(ii)} a set of admissible (ungrounded) fluents \(\mathcal{F}\) (i.e., all the fluents present in the PDDL files in general form, e.g., \textit{(robot-at ?var1)}); \emph{(iii)} a set of admissible objects (i.e., all the objects present in the problem file, e.g., \textit{loc0}); and outputs  a set of (grounded) fluents $\hat{f}$ derived from the state description (e.g., \textit{(robot-at loc0)}).\looseness-1

Let $f_t$ represent the actual set of fluents describing a PDDL state at time $t$ and $\hat{f}$ the set of fluents predicted by the robot answer. 
We measure the soundness $S_t$ and completeness $C_t$ of the response with respect to the state of the plan at time $t$ as:
\vspace{-0.5em}
\begin{equation}
S_t =\frac{|\hat{f} \cap f_t|}{|\hat{f}|} \quad C_t =\frac{|\hat{f} \cap f_t|}{|f_t|}  
\end{equation}
Note that soundness is maximized (equal to 1) when the response contains \textit{only} fluents that are actually true in the state of the plan at time $t$, while completeness $C_t$ is maximized when the response predicts \textit{all} the fluents that are true in the state of the plan at time $t$.\looseness-1


Finally, we define the soundness and completeness of statements concerning the present, past, and future as:
\vspace{-0.5em}
\begin{equation}
    S_{present} = S_t \quad C_{present} = C_t
\end{equation}
\vspace{-1em}
\begin{equation}
S_{past} = \frac{1}{t} \sum_{i=1}^{t}S_i 
\quad C_{past} = \frac{1}{t} \sum_{i=1}^{t}C_i 
\end{equation}
\begin{equation}
S_{future} = \frac{1}{T-t+1} \sum_{i=t}^{T} S_i
\quad C_{future} = \frac{1}{T-t+1} \sum_{i=t}^{T} C_i
\end{equation}
With $t$ being the current time and $T$ the final time, and $1\leq t< T$ for the past, and $1 < t \leq T$ for the future.

It is important to note that the planner operates offline, meaning the policy is computed before execution begins. Thus, assuming knowledge of the policy in advance is a valid assumption, particularly in the case of evaluating future states. Additionally, to compute the metrics about the future, we have to \textit{determinize} the policy by removing all \textit{non-deterministic} decision nodes, grounding them with one of the possible events that can occur.

\subsection{Experiments}

\subsubsection{Process Extraction}
We assessed the accuracy of Lang2LTL on sentences generated for the \projectname{} domain through three distinct experiments:

\noindent \textbf{Vanilla Lang2LTL}: In this test, we use Lang2LTL with: \emph{(i)} its pre-trained symbolic model, \emph{(ii)} its standard RER prompt that includes 16 examples, and \emph{(iii)} our specific landmark file tailored to the domain.\looseness-1
    
\noindent \textbf{Lang2LTL with augmented RER prompt (11 symbols)}: For this experiment, we again employ Lang2LTL using: \emph{(i)} the pre-trained symbolic model, \emph{(ii)} an enhanced RER prompt containing 34 examples, which includes 18 instances of RER in non-navigational generic task sentences, and \emph{(iii)} our landmark file. Notably, this RER prompt uses sentences not in the test set and incorporates only 11 of the 23 symbols in our domain (including 4 location symbols, 4 condition symbols, and 3 action symbols).\looseness-1
    
\noindent \textbf{Lang2LTL with augmented RER prompt (18 symbols)}: This experiment closely mirrors the previous one, but we further support Lang2LTL in the RER phase with an enhanced prompt that still consists of 18 sentences. This time, it encompasses nearly all symbols in our domain—18 out of 23 (the 4 locations, 8 conditions, and all 6 actions).\looseness-1

\begin{table}[t]
  \centering
  \caption{Accuracy of Lang2LTL on navigation and generic tasks}
    \begin{tabular}{@{}lll@{}}
    \toprule
    \textbf{Experiment} & \textbf{Navigation tasks} & \textbf{Generic tasks} \\ \midrule
    Vanilla Lang2LTL             & 86\%                      & 22\%                   \\
    RER-prompt 1        & \textbf{86\%}                      & 77\%                   \\
    RER prompt 2        & 85\%                      & \textbf{88\%}                   \\ \bottomrule
    \end{tabular}%
  \label{tab:lang2ltl-accuracy}%
\end{table}%

    \begin{table}[t]
\caption{Evaluation of the robot answers in the $3$ considered settings}
\label{llm_table}
\begin{center}
\begin{tabular}{@{}lll@{}}
\toprule
\textbf{Category} & \textbf{Soundness} & \textbf{Completeness}\\
\midrule
\textit{What are you doing now?} & \textbf{0.99 $\pm$ 0.03} & $0.67 \pm 0.11$\\
\textit{What did you do so far?} & $0.96 \pm 0.04$ & \textbf{0.80 $\pm$ 0.08}\\
\textit{What are you going to do next?} & $0.88 \pm 0.03$ & $0.58 \pm 0.08$\\
\bottomrule
\end{tabular}
\vspace{-15pt}
\end{center}
\end{table}

The results are presented in Table \ref{tab:lang2ltl-accuracy}.
As shown, Vanilla Lang2LTL exhibits poor performance on generic tasks. This shortfall is primarily due to the system's difficulty in recognizing expressions in sentences that refer to actions or conditions. In contrast, enhancing the RER prompt leads to a notable improvement in performance for generic tasks, while navigation tasks remain stable. The assistance provided to the tool in identifying referring expressions correlates positively with performance. The best outcomes arise from using the RER prompt with examples covering 18 out of 23 symbols, surpassing the highest results achieved in navigation tasks.

Upon reviewing the incorrect predictions from the last experiment, we discovered that, aside from a few grounding errors, many formulas were incorrect even when the RER and grounding were accurate. Thus, achieving 88\% accuracy appears to be the upper limit for generic tasks without retraining the symbolic translation module, relying solely on optimizing the landmark file and RER prompt. Nonetheless, the symbolic module has demonstrated strong robustness, effectively handling non-navigational tasks when adequately supported by the other modules.\looseness-1

\subsubsection{Process Monitoring}

To assess the ability of the robot to monitor the process, we conducted 30 experiments for each of the three cases, totaling 90 experiments. In each experiment we set \textit{temperature} for both the LLM and the $FluentExtractor$, to $1e-7$ to minimize randomness. 
The human user's question could arise at any point during the policy execution—whether immediately after the robot begins deliberating, just before it reaches the goal, or at any moment in between. To simulate this behavior, we query the LLM after each action execution in the determinized plan, with a probability $p$ that increases after each action. This approach ensures that a question will eventually be posed before the execution concludes. \looseness-1

\begin{figure}[t!]
    \centering
    \includegraphics[width=\linewidth]{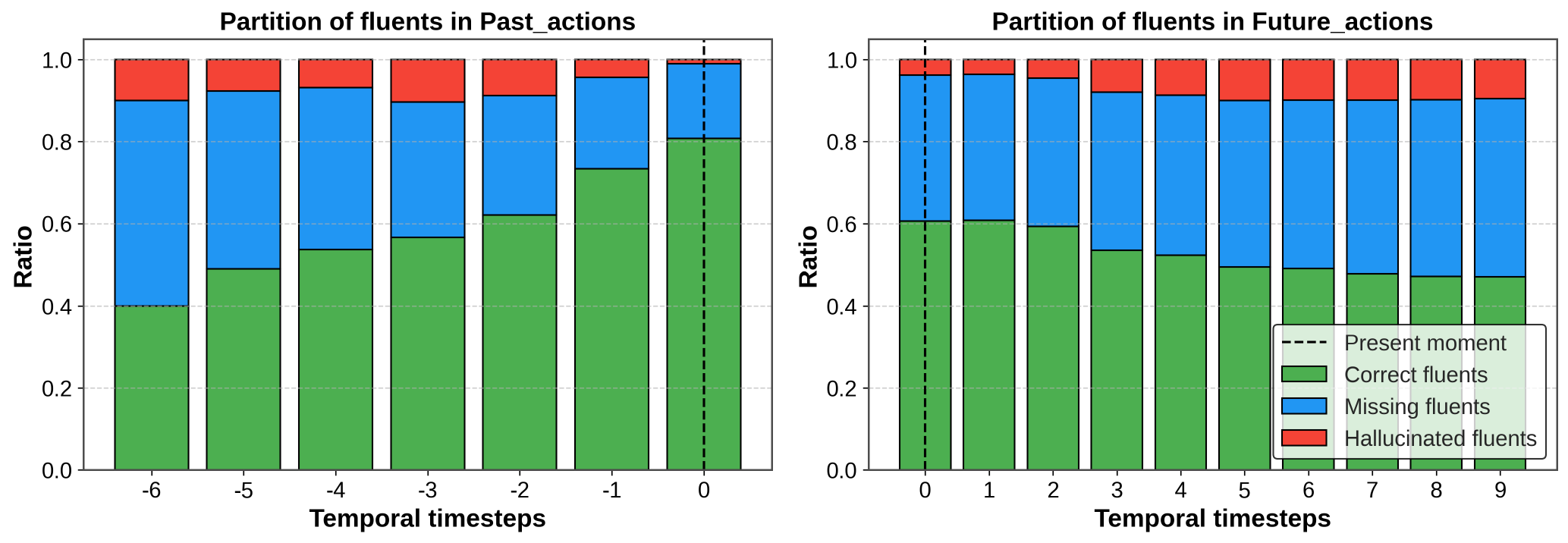}
    \caption{Ratio of correct (in green), incorrect (red), and missing (blue) fluent predictions when evaluated against past (left) and future (right) states.}
    \vspace{-15pt}
    \label{fig:histograms}
\end{figure}

\begin{figure*}[t!]
    \centering
    \includegraphics[width=\linewidth]{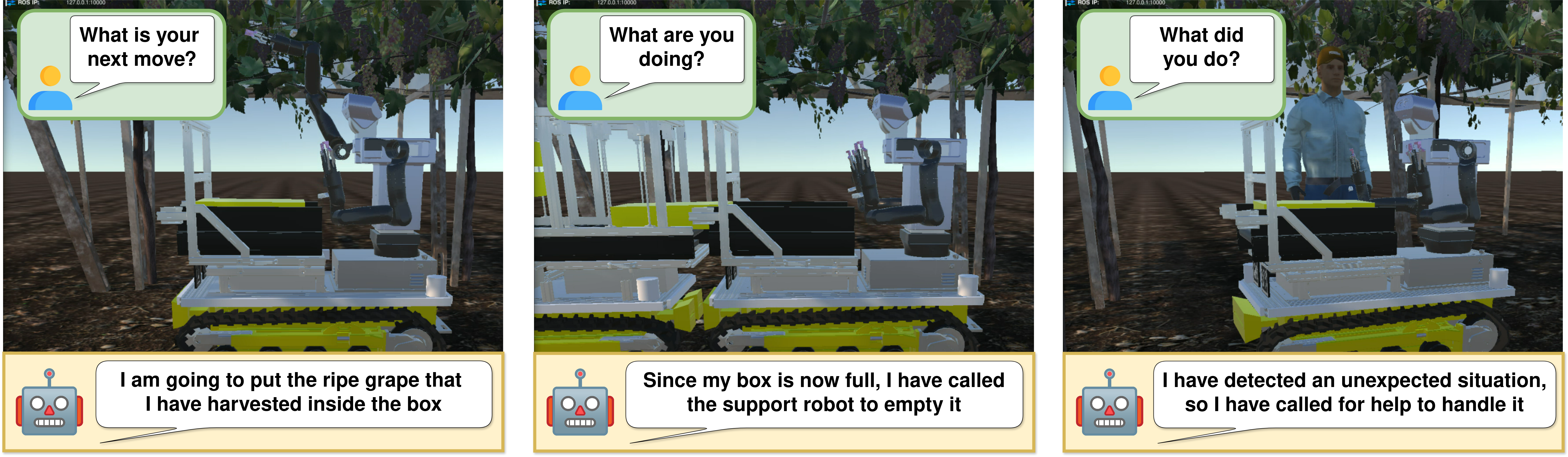}
    \caption{A display of possible interactions inside the simulated environment of \projectname{}. On the left, the user is asking the robot for its next intentions. In the middle, the user is making questions about the current situation. On the right, the user is asking about past actions of the robot.}
    \label{fig:simulator}
    \vspace{-15pt}
\end{figure*}

We report soundness and completeness for the three scenarios in Table~\ref{llm_table}. Note that soundness is very close to one, especially for responses concerning the present and the past. This indicates that the responses are almost \textit{entirely free of hallucinations}, which is a crucial characteristic when using LLMs.
Conversely, completeness is lower than its maximum value but still surprisingly high. Achieving a completeness of one would imply that the robot's statements allow for the full reconstruction of all fluents that are true in the planning domain at the present time ($C_{present}$), at every past instant ($C_{past}$), or at every future instant ($C_{future}$). However, it is reasonable that this is not the case. The responses naturally focus on the most relevant domain information needed to justify decisions while omitting less critical details.

In Figure \ref{fig:histograms}, we illustrate the percentage of fluents that are: \emph{(i)} correctly predicted (belonging to $f_t \cap \hat{f}$), \emph{(ii)} hallucinated (predicted but not actually part of the state), and \emph{(iii)} missing (part of the state but not included in the prediction) across different time instants for statements about the past and future.
From the figure, it is evident that responses concerning the past and future tend to describe nearby time instants more accurately, with quality degrading as the temporal distance from the present increases. This is reasonable, as that the plan can be long, and it is easier to respond about recently occurred states or those in the near future.\looseness-1

\subsubsection{Qualitative Evaluation}

A qualitative evaluation was conducted by instantiating an environment that respects the PDDL specifications, inside the official \projectname{}' Unity-based simulator (see Figure \ref{fig:simulator}). The simulator represents one-to-one the expected robot behavior in the vineyard when collaborating with human farmers for grape harvesting. Inside the simulated environment, the robot takes as input the fully generated plan that achieves the LTLf goal, and executes it action by action, showing how the environment changes when the robot is deliberating. At any time during the plan execution, it is possible to interrupt the robot and question it about what it is doing, what it has done, or what it will do. The evaluation was conducted by involving seven expert farmers. The answers provided confirmed the results obtained with the synthetic dataset. In addition, we observed that statements about the present generally do not describe only the present but place it in the context of the immediate past and future, often including references to activities performed in the previous step or those intended to be performed in the next step (see Figure \ref{fig:simulator} (middle)). A full video demonstration is available on the paper website.


\section{CONCLUSIONS AND FUTURE WORKS}
\label{sec:conclusions}

In this paper, we introduced a framework that combines LLMs and automated planning to let human users specify a high-level robot activity and monitor its execution. As discussed in Section~\ref{sec:beyond}, the approach can be adapted to different levels of granularity, depending on the actions implemented on the robotic platform. We evaluated it on a synthetic dataset for precision agriculture.

For our experiments, we used GPT. While this model cannot be run locally, we observe that: \emph{(i)} the number of calls needed for a single process description is limited; \emph{(ii)} the methodology is general enough that GPT-4 could be replaced by any current or future LLM; and \emph{(iii)} with ongoing advances in hardware and open-source models, effective performance should soon be achievable with offline LLMs.\looseness-1

Despite the positive results, human queries may require reasoning beyond current LLM capabilities. An emerging direction develops LLMs as tool-using agents, enabling complex problem solving and reducing hallucinations by planning calls to external services. Such tools can access external information (e.g., documents, calendars) or act in virtual and physical settings (e.g., robotic arms). The strategy leverages in-context learning by embedding API capabilities and invocation patterns in the prompt. While this mitigates hallucinations, it constrains responses to tool outputs; therefore, our tool-free approach may be preferable from a human-operator perspective.\looseness-1

Regarding process extraction, the component currently handles specific patterns of LTL formulas. Extracting generic LTL from text remains open, though research is progressing rapidly.\looseness-1

Our evaluation focuses on quantitative monitoring accuracy; future work includes a user study with domain experts.\looseness-1

Finally, the proposed solution does not monitor the human while engaged in the process. LLMs could also ingest natural language reports of completed actions to complement the robot’s knowledge of the high-level activity status.

\section*{Acknowledgment}
This work has been carried out while Francesco Argenziano was enrolled in the Italian National Doctorate on Artificial Intelligence run by Sapienza University of Rome. The work of Francesco Leotta has been partly funded by the Sapienza project “Cognitive and observable cloud continuum” (Proposal ID 101092792). This work has been partially supported by the EU Horizon 2020 research and innovation programme under grant agreement No 101016906 – Project CANOPIES. We acknowledge partial financial support from PNRR MUR project PE0000013-FAIR.

\bibliographystyle{IEEEtran}
\bibliography{mybibfile}

\end{document}